\documentclass[11pt,a4paper]{article}
\usepackage[hyperref]{acl2021}
\usepackage{times}
\usepackage{latexsym}

\usepackage{anyfontsize}
\usepackage{t1enc}
\usepackage{graphicx}
\usepackage[export]{adjustbox}
\usepackage{caption}
\usepackage{subcaption}
\usepackage{tipa}
\usepackage{multirow}
\usepackage{rotating} 
\newcommand{\minitab}[2][l]{\begin{tabular}{#1}#2\end{tabular}}

\usepackage{microtype}

\aclfinalcopy 
\def\aclpaperid{1309} 

\title{Exploring the Representation of Word Meanings in Context: A Case Study on Homonymy and Synonymy}

\author{Marcos Garcia \\
  CiTIUS -- Research Center in Intelligent Technologies \\
  Universidade de Santiago de Compostela, Galiza \\
  \texttt{marcos.garcia.gonzalez@usc.gal}}%

\date{}

\begin{document}
\maketitle
\begin{abstract}
This paper presents a multilingual study of word meaning representations in context. We assess the ability of both static and contextualized models to adequately represent different lexical-semantic relations, such as homonymy and synonymy. To do so, we created a new multilingual dataset that allows us to perform a controlled evaluation of several factors such as the impact of the surrounding context or the overlap between words, conveying the same or different senses.
A systematic assessment on four scenarios shows that the best monolingual models based on Transformers can adequately disambiguate homonyms in context. However, as they rely heavily on context, these models fail at representing words with different senses when occurring in similar sentences.
Experiments are performed in Galician, Portuguese, English, and Spanish, and both the dataset (with more than 3,000 evaluation items) and new models are freely released with this study.
\end{abstract}

\section{Introduction}
Contrary to static vector models, which represent the different senses of a word in a single vector \citep{erk2012vector,mikolov2013}, contextualized models generate representations at token-level \citep{peters-etal-2018-deep,devlin-etal-2019-bert}, thus being an interesting approach to model word meaning in context. In this regard, several studies have shown that clusters produced by some contextualized word embeddings (CWEs) are related to different senses of the same word \cite{NEURIPS2019_159c1ffe,wiedemann-etal-2019-does}, 
or that similar senses can be aligned in cross-lingual experiments \cite{schuster-etal-2019-cross}.

However, more systematic evaluations of polysemy (i.e., word forms that have different related meanings depending on the context 
\citep{apresjan1974regular}), have shown that even though CWEs present some correlations with human judgments \citep{nair-etal-2020-contextualized}, they fail to predict the similarity of the various senses of a polysemous word \citep{haber-poesio-2020-assessing}.

As classical datasets to evaluate the capabilities of vector representations consist of single words without context \citep{finkelstein2001placing} or heavily constrained expressions \citep{kintsch2001predication,mitchell-lapata-2008-vector}, new resources with annotations of words in free contexts have been created, including both graded similarities \citep{huang-etal-2012-improving,armendariz-etal-2020-cosimlex} or binary classification of word senses \citep{pilehvar-camacho-collados-2019-wic,raganato-etal-2020-xl}. However, as these datasets largely include instances of polysemy, they are difficult to solve even for humans (in fact, the highest reported human upper bound is about 80\%) as the nuances between different senses depend on non-linguistic factors such as the annotator procedure or the target task \citep{tuggy1993ambiguity,kilgarriff1997don,hanks2000,erk-2010-word}.

In this paper, we rely on a more objective and simple task to assess how contextualized approaches (both neural network models and contextualized methods of distributional semantics) represent word meanings in context. In particular, we observe whether vector models can identify unrelated meanings represented by the same word form (homonymy) and the same sense conveyed by different words (synonymy). In contrast to polysemy, there is a strong consensus concerning the representation of homonymous senses in the lexicon, and it has been shown that homonyms are cognitively processed differently than polysemous words \citep{klepousniotou2012not,macgregor2015sustained}. In this regard, exploratory experiments in English suggest that some CWEs correctly model homonymy, approximating the contextualized vectors of a homonym to those of its paraphrases \citep{lake2020word}, and showing stronger correlation with human judgments to those of polysemous words \citep{nair-etal-2020-contextualized}. However, as homonyms convey unrelated meanings depending on the context, it is not clear whether the good performance of CWEs actually derives from the contextualization process or simply from the use of explicit lexical cues present in the sentences.

Taking the above into account, we have created a new multilingual dataset (in Galician, Portuguese, English, and Spanish) with more than 3,000 evaluation items. It allows for carrying out more than 10 experiments and controlling factors such as the surrounding context, the word overlap, and the sense conveyed by different word forms.
We use this resource to perform a systematic evaluation of contextualized word meaning representations. We compare different strategies using both static embeddings and current models based on deep artificial neural networks. The results suggest that the best monolingual models based on Transformers \citep{vaswani2017attention} can identify homonyms having different meanings adequately. However, as they strongly rely on the surrounding context, words with different meanings are represented very closely when they occur in similar sentences. Apart from the empirical conclusions and the dataset, this paper also contributes with new BERT and \textit{fastText} models for Galician.\footnote{Dataset, models, and code are available at \url{https://github.com/marcospln/homonymy_acl21/}.} 

Section~\ref{sec:rw} presents previous studies about word meaning representation. Then, Section~\ref{sec:dataset} introduces the new dataset used in this paper. In Section~\ref{sec:models} we describe the models and methods to obtain the vector representations. Finally, the experiments and results are discussed in Section~\ref{sec:eval}, while Section~\ref{sec:conclusions} draws some conclusions of our study.

\section{Related Work}
\label{sec:rw}
A variety of approaches has been implemented to compute word meaning in context by means of standard methods of distributional semantics \citep{schutze-1998-automatic,kintsch2001predication,mcdonald-brew-2004-distributional,erk-pado-2008-structured}. As compositional distributional models construct sentence representations from their constituents vectors, they take into account contextualization effects on meaning \citep{mitchell-lapata-2008-vector,baroni-zamparelli-2010-nouns,baroni2013composition}. However, these approaches often have scalability problems as their representations grow exponentially with the size of the sentences. Therefore, the datasets used to evaluate them are composed of highly restricted phrases \citep{grefenstette-sadrzadeh-2011-experimental}.

The rise of artificial neural networks on natural language processing popularized the use of vector representations, and the remarkable performance of neural language models \citep{melamud-etal-2016-context2vec,peters-etal-2018-deep} led to a productive line of research exploring to what extent these models represent linguistic knowledge \citep{bertology}. However, few of these works have focused on lexical semantics, and most of the relevant results in this field come from evaluations in downstream tasks. In this regard, \citet{wiedemann-etal-2019-does} found that clusters of BERT embeddings \cite{devlin-etal-2019-bert} seem to be related to word senses, while \citet{schuster-etal-2019-cross} observed that clusters of polysemous words correspond to different senses in a cross-lingual alignment of vector representations.

Probing LSTMs on lexical substitution tasks, \citet{aina-etal-2019-putting} showed that these architectures rely on the lexical information from the input embeddings, and that the hidden states are biased towards contextual information. On an exploration of the geometric representations of BERT, \citet{NEURIPS2019_159c1ffe} found that different senses of a word tend to appear separated in the vector space, while several clusters seem to correspond to similar senses.
Recently, \citet{vulic-etal-2020-probing} evaluated the performance of BERT models on several lexical-semantic tasks in various languages, including semantic similarity or word analogy. The results show that using special tokens ([CLS] or [SEP]) hurts the quality of the representations, and that these tend to improve across layers until saturation. As this study uses datasets of single words (without context), type-level representations are obtained by averaging the contextualized vectors over various sentences.

There are several resources to evaluate word meaning in free contexts, such as the Stanford Contextual Word Similarity \citep{huang-etal-2012-improving} and CoSimLex \citep{armendariz-etal-2020-cosimlex}, both representing word similarity on a graded scale, or the Word-in-Context datasets (WiC), focused on binary classifications (i.e., each evaluation item contains two sentences with the same word form, having the same or different senses) \citep{pilehvar-camacho-collados-2019-wic,raganato-etal-2020-xl}. These datasets include not only instances of homonymy but mostly of polysemous words. In this regard, studies on polysemy using Transformers have obtained diverse results: \citet{haber-poesio-2020-assessing} found that BERT embeddings correlate better with human ratings of co-predication than with similarity between word senses, thus suggesting that these representations encode more contextual information than word sense knowledge. Nevertheless, the results of \citet{nair-etal-2020-contextualized} indicate that BERT representations are correlated with human scores of polysemy. An exploratory experiment of the latter study also shows that BERT discriminates between polysemy and homonymy, which is also suggested by other pilot evaluations reported by \citet{lake2020word} and \citet{yu-ettinger-2020-assessing}.

Our study follows this research line pursuing objective and unambiguous lexical criteria such as the representation of homonyms and synonyms.
In this context, there is a broad consensus in the psycholinguistics literature regarding the representation of homonyms as different entries in the lexicon (in contrast to polysemy, for which there is a long discussion on whether senses of polysemous words are stored as a single core representation or as independent entries \citep{hogeweg2020nature}). In fact, several studies have shown that homonyms are cognitively processed differently from polysemous words \citep{klepousniotou2012not,rabagliati2013truth}. In contrast to the different senses of polysemous words, which are simultaneously activated, the meanings of homonyms are in conflict during processing, with the not relevant ones being deactivated by the context \citep{macgregor2015sustained}.
To analyze how vector models represent homonymy and synonymy in context, we have built a new multilingual resource with a strong inter-annotator agreement, presented below.

\section{A New Multilingual Resource of Homonymy and Synonymy in Context}
\label{sec:dataset}
This section briefly describes some aspects of lexical semantics relevant to our study, and then presents the new dataset used in the paper.

\paragraph{Homonymy and homography:}
Homonymy is a well-known type of lexical ambiguity that can be described as the relation between distinct and unrelated meanings represented by the same word form, such as \textit{match}, meaning for instance `sports game' or `stick for lighting fire'. In contrast to polysemy (where one lexeme conveys different related senses depending on the context, e.g., \textit{newspaper} as an organization or as a set of printed pages), it is often assumed that homonyms are different lexemes that have the same lexical form \citep{cruse1986lexical}, and therefore they are stored as independent entries in the lexicon \citep{pustejovsky1998generative}.

There are two main criteria for homonymy identification: Diachronically, homonyms are lexical items that have different etymologies but are accidentally represented by the same word form, while a synchronic perspective strengthens unrelatedness in meaning. Even if both approaches tend to identify similar sets of homonyms, there may be ambiguous cases that are diachronically but not synchronically related (e.g., two meanings of \textit{banco} --`bench' and 
`financial institution'-- in Portuguese or Spanish could be considered polysemous as they derive from the same origin,\footnote{In fact, several dictionaries organize them in a single entry: \url{https://dicionario.priberam.org/banco},\\\url{https://dle.rae.es/banco}.} but as this is a purely historical association, most speakers are not aware of the common origin of both senses). In this study, we follow the synchronic perspective, and consider homonymous meanings those that are clearly unrelated (e.g., they unambiguously refer to completely different concepts) regardless of their origin.

It is worth mentioning that as we are dealing with written text we are actually analyzing homographs (different lexemes with the same spelling) instead of homonyms. Thus, we discard instances of phonologically identical words which are written differently, such as the Spanish \textit{hola} `hello' and \textit{ola} `wave', both representing the phonological form \textipa{/ola/}. Similarly, we include words with the same spelling representing different phonological forms, e.g., the Galician-Portuguese \textit{sede}, which corresponds to both \textipa{/sede/} `thirst', and \textipa{/sEde/} `headquarters'.

In this paper, \textit{homonymous senses} are those unrelated meanings conveyed by the same (homonym) word form. For instance, \textit{coach} may have two homonymous senses (`bus' and `trainer'), which can be conveyed by other words (synonyms) in different contexts (e.g., by \textit{bus} or \textit{trainer}).

\begin{table*}[!ht]
\centering
\begin{tabular}{|c|lll|}
\hline \textbf{Sense} & \textbf{Sentences 1-3} & \textbf{Sentence 4} & \textbf{Sentence 5}\\ \hline
\multirow{3}{*}{(1)} & \small{We're going to the airport by \textbf{coach}}. & \multirow{3}{*}{\minitab[l]{\small{[\ldots] the \textbf{coach} was badly}\\\small{delayed by roadworks.}}} & \multirow{3}{*}{\minitab[l]{\small{They had to travel}\\\small{everywhere by \textbf{bus}.}}}\\
& \small{We're going to the airport by \textbf{bus}.} & & \\
& \small{We're going to the airport by \textit{bicycle}.} & &\\ \hline

\multirow{3}{*}{(2)} & \small{That man was appointed as the new \textbf{coach}.} & \multirow{3}{*}{\minitab[l]{\small{She has recently joined}\\\small{the amateur team as \textbf{coach}.}}} & \multirow{3}{*}{\minitab[l]{\small{They need a new \textbf{trainer}}\\\small{for the young athletes.}}}\\
& \small{That man was appointed as the new \textbf{trainer}.} & & \\
& \small{That man was appointed as the new \textit{president}.} & &\\
\hline
\end{tabular}
\caption{\label{tab:sentences} Example sentences for two senses of \textit{coach} in English (`bus' and `trainer'). Sentences 1 to 3 include, in the same context, the target word, a synonym, and a word with a different sense (in italic), respectively. Sentences 4 and 5 contain the target word and a synonym in different contexts, respectively.}
\end{table*}

\paragraph{Structure of the dataset:}
We have created a new resource to investigate how vector models represent word meanings in context. In particular, we want to observe whether they capture (i) different senses conveyed by the same word form (homonymy), and (ii) equivalent senses expressed by different words (synonymy). The resource contains controlled sentences so that it allows us to observe how the context and word overlap affect word representations.

To allow for different comparisons with the same and different contexts, we have included five sentences for each meaning (see Table~\ref{tab:sentences} for examples): three sentences containing the target word, a synonym, and a word with a different sense, all of them in the same context (sentences 1 to 3), and two additional sentences with the target word and a synonym, representing the same sense (sentences 4 and 5, respectively). Thus, for each sense we have four sentences (1, 2, 4, 5) with a word conveying the same sense (both in the same and in different contexts) and another sentence (3) with a different word in the same context as sentences 1 and 2.

From this structure, we can create datasets of sentence triples, where the target words of two of them convey the same sense, and the third one has a different meaning. Thus, we can generate up to 48 triples for each pair of senses (24 in each direction: sense 1 vs. sense 2, and vice-versa). These datasets allow us to evaluate several semantic relations at the lexical level, including homonymy, synonymy, and various combinations of homonymous senses. Interestingly, we can control for the impact of the context (e.g., are contextualized models able to distinguish between different senses occurring in the same context, or do they incorporate excessive contextual information into the word vectors?), the word overlap (e.g., can a model identify different senses of the same word form depending on the context, or it strongly depends on lexical cues?), or the POS-tag (e.g., are homonyms with different POS-tags easily disambiguated?). 

\begin{table*}[!ht]
\centering
\begin{tabular}{|c|rrr|rrr|c|}
\hline \textbf{Language} & \textbf{Hom.} & \textbf{Senses} & \textbf{Sent.} & \textbf{Triples} & \textbf{Pairs} & \textbf{WiC} & $\kappa$\\ \hline
Galician & 22 & 47 (4) & 227 & 1365 & 823 & 197 & 0.94 \\
English & 14 & 30 (5) & 138 & 709 & 463 & 129 & 0.96 \\
Portuguese & 11 & 22 (1) & 94 & 358 & 273 & 81 & 0.96 \\
Spanish & 10 & 23 (3) & 105 & 645 & 391 & 101 & 0.95 \\
\hline
Total & 57 & 122\phantom{(0)} & 564 & 3077 & 1950 & 508 & 0.94 \\
\hline
\end{tabular}
\caption{\label{tab:size} Characteristics of the dataset. First three columns display the number of homonyms (\textit{Hom}), senses, and sentences (\textit{Sent}), respectively. Senses in parentheses are the number of homonymous pairs with different POS-tags). Center columns show the size of the evaluation data in three formats: triples, pairs, and WiC-like pairs, followed by the Cohen's $\kappa$ agreements and their micro-average. The total number of homonyms and senses is the sum of the language-specific ones, regardless of the fact that some senses occur in more than one language.}
\end{table*}

\paragraph{Construction of the dataset:}
We compiled data for four languages: Galician, Portuguese, Spanish, and English.\footnote{Galician is generally considered a variety of a single (Galician-)Portuguese language. However, they are divided in this resource, as Galician has recently been standardized using a Spanish-based orthography that formally separates it from Portuguese \cite{samartim2012}.} We tried to select sentences compatible with the different varieties of the same language (e.g., with the same meaning in UK and US English, or in Castilian and Mexican Spanish). However, we gave priority to the European varieties when necessary (e.g., regarding spelling variants).

The dataset was built using the following procedure: First, language experts (one per language) compiled lists of homonyms using dedicated resources for language learning, together with WordNet and other lexicographic data \cite{miller1995wordnet,montraveta2010construccion,guinovart2011galnet,rademaker-etal-2014-openwordnet}. Only clear and unambiguous homonyms were retained (i.e., those in the extreme of the \textit{homonymy-polysemy-vagueness} scale \cite{tuggy1993ambiguity}). These homonyms were then enriched with frequency data from large corpora: Wikipedia and SLI GalWeb \cite{agerri-etal-2018-developing} for Galician, and a combination of Wikipedia and Europarl for English, Spanish and Portuguese \cite{koehn2005europarl}. From these lists, each linguist selected the most frequent homonyms, annotating them as ambiguous at type or token level (\textit{absolute homonymy} and \textit{partial homonymy} in Lyons' terms \cite{lyons1995linguistic}). As a substantial part were noun-verb pairs, only a few of these were included. For each homonym, the language experts selected from corpora two sentences (1 and 4) in which the target words were not ambiguous.\footnote{Sentences were selected, adapted, and simplified using GDEX-inspired constraints \cite{kilgarriff2008gdex} (i.e., avoiding high punctuation ratios, unnecessary subordinate clauses, etc.), which resulted in the creation of new sentences.} They then selected a synonym that could be used in sentence 1 without compromising grammaticality (thus generating sentence 2), and compiled an additional sentence for it (5), trying to avoid further lexical ambiguities in this process.\footnote{In most cases, this synonym is the same as that of sentence 2, but this is not always the case. Besides, in some cases we could not find words conveying the same sense, for which we do not have sentences 2 and 5.} For each homonym, the linguists selected a word with a different meaning (for sentence 3), trying to maximize the following criteria: (i) to refer unambiguously to a different concept, and to preserve (ii) semantic felicity and (iii) grammaticality. The size of the final datasets varies depending on the initial lists and on the ease of finding synonyms in context. 

\paragraph{Results:}
Apart from the sentence triples explained above, the dataset structure allows us to create evaluation sets with different formats, such as sentence pairs to perform binary classifications as in the WiC datasets. Table~\ref{tab:size} shows the number of homonyms, senses, and sentences of the multilingual resource, together with the size of the evaluation datasets in different formats.

As the original resource was created by one annotator per language, we ensured its quality as follows: We randomly extracted sets of 50 sentence pairs and gave them to other annotators (5 for Galician, and 1 for each of the other three varieties, all of them native speakers of the target language). We then computed the Cohen's $\kappa$ inter-annotator agreement \citep{cohen1960coefficient} between the original resource and the outcome of this second annotation (see the right column of Table~\ref{tab:size}). We obtained a micro-average $\kappa=0.94$ across languages, a result which supports the task's objectivity. Nevertheless, it is worth noting that few sentences have been carefully modified after this analysis, as it has shown that several misclassifications were due to the use of an ambiguous synonym. Thus, it is likely that the final resource has higher agreement values.

\section{Models and Methods}
\label{sec:models}
This section introduces the models and procedures to obtain vector representations followed by the evaluation method.

\begin{table*}[!ht]
\centering
\begin{tabular}{|l|rr|rr|rr|rr|rr||rr|}
\hline
{\bf Language} & \multicolumn{2}{c|}{\bf Exp1} & \multicolumn{2}{c|}{\bf Exp2} & \multicolumn{2}{c|}{\bf Exp3} & \multicolumn{2}{c|}{\bf Exp4} & \multicolumn{2}{c||}{\bf Total} & \multicolumn{2}{c|}{\bf Full}\\ \hline
Galician & 122 & 105 & 183 & 149 & 278 & 229 & 135 & 135 & 718 & 618 & 1365 & 1157 \\
English & 77 & 52 & 89 & 58 & 144 & 91 & 68 & 68 & 378 & 269 & 709 & 494 \\
Portuguese & 45 & 41 & 37 & 37 & 80 & 74 & 41 & 41 & 203 & 193 & 358 & 342 \\
Spanish & 65 & 49 & 87 & 71 & 146 & 110 & 59 & 59 & 357 & 289 & 645 & 517 \\
\hline
\end{tabular}
\caption{\label{tab:eval_triples}Number of instances of each experiment and language. Numbers on the right of each column are those triples where the three target words belong to the same morphosyntactic category (left values are the total number of triples). \textit{Total} are the sums of the four experiments, while \textit{Full} refers to all the instances of the dataset.}
\end{table*}

\subsection{Models}
We have used static embeddings and CWEs based on Transformers, comparing different ways of obtaining the vector representations in both cases:

\paragraph{Static embeddings:}
We have used skip-gram \textit{fastText} models of 300 dimensions \citep{bojanowski-etal-2017-enriching}.\footnote{In preliminary experiments we also used \textit{word2vec} and GloVe models, obtaining slightly lower results than \textit{fastText}.} For English and Spanish, we have used the official vectors trained on Wikipedia. For Portuguese, we have used the model provided by \citet{hartmann-etal-2017-portuguese}, and for Galician we have trained a new model (see Appendix~\ref{app:models} for details).\footnote{These Portuguese and Galician models obtained better results (0.06 on average) than the official ones.}

\paragraph{Contextualized embeddings:} We have evaluated multilingual and monolingual models:\footnote{To make a fair comparison we prioritized \textit{base} models (12 layers), but we also report results for \textit{large} (24 layers) and 6 layers models when available.}

\textbf{Multilingual models:} We have used the official multilingual BERT (mBERT cased, 12 layers) \citep{devlin-etal-2019-bert}, XLM-RoBERTa (Base, 12 layers) \citep{conneau-etal-2020-unsupervised}, and DistilBERT (DistilmBERT, 6 layers) \citep{sanh2019distilbert}.

\textbf{Monolingual models:} For English, we have used the official BERT-Base model (uncased). For Portuguese and Spanish, BERTimbau \citep{souza2020bertimbau} and BETO \citep{CaneteCFP2020} (both cased). For Galician, we trained two BERT models (with 6 and 12 layers; see Appendix~\ref{app:models}).

\subsection{Obtaining the vectors}
\paragraph{Static models:} These are the methods used to obtain the representations from the static models:

\textbf{Word vector (\textit{WV}):} Embedding of the target word (homonymous senses with the same word form will have the same representation).

\textbf{Sentence vector (\textit{Sent}):} Average embedding of the whole sentence.

\textbf{Syntax (\textit{Syn}):} Up to four different representations obtained by adding the vector of the target word to those of their syntactic heads and dependents. This method is based
on the assumption that the syntactic context of a word characterizes its meaning, providing relevant information for its contextualized representation (e.g., in `He swims to the bank', \textit{bank} may be disambiguated by combining its vector with the one of \textit{swim}).\footnote{We have also evaluated a contextualization method using selectional preferences inspired by \citet{erk-pado-2008-structured}, but the results were almost identical to those of the \textit{WV} approach.} Appendix~\ref{app:syn} describes how heads and dependents are selected.

\paragraph{Contextualized models:}
For these models, we have evaluated the following approaches:

\textbf{Sentence vector (\textit{Sent}):} Vector of the sentence built by averaging all words (except for the special tokens [CLS] and [SEP]), each of them represented by the standard approach of concatenating the last 4 layers \cite{devlin-etal-2019-bert}.

\textbf{Word vector (\textit{WV}):} Embedding of the target word, combining the vectors of the last 4 layers. We have evaluated two operations: vector concatenation (\textit{Cat}), and addition (\textit{Sum}).

\textbf{Word vector across layers (\textit{Lay}):} Vector of the target word on each layer. This method allows us to explore the contextualization effects on each layer.

Vectors of words split into several sub-words are obtained by averaging the embeddings of their components. Similarly, MWEs vectors are the average of the individual vectors of their components, both for static and for contextualized embeddings.

\subsection{Measuring sense similarities}
Given a sentence triple where two of the target words (\textit{a} and \textit{b}) have the same sense and the third (\textit{c}) a different one, we evaluate a model as follows (in a similar way as other studies \citep{kintsch2001predication,lake2020word}): First, we obtain three cosine similarities between the vector representations: $sim1=cos(a, b)$; $sim2=cos(a, c)$; $sim3=cos(b, c)$. Then, an instance is labeled as \textit{correct} if those words conveying the same sense (\textit{a} and \textit{b}) are closer together than the third one (\textit{c}). In other words, $sim1>sim2$ and $sim1>sim3$: Otherwise, the instance is considered as \textit{incorrect}.

\section{Evaluation}
\label{sec:eval}
This section presents the experiments performed using the new dataset and discusses their results.

\subsection{Experiments}
Among all the potential analyses of our data, we have selected four evaluations to assess the behavior of a model by controlling factors such as the context and the word overlap:

\paragraph{Homonymy (Exp1):} The same word form in three different contexts, two of them with the same sense (e.g., \textit{coach} in sentences [1:1, 1:4, 2:1]\footnote{First and second digits refer to the sense and sentence ids.} in Table~\ref{tab:sentences}). This test evaluates if a model correctly captures the sense of a unique word form in context. \noindent\textbf{Hypothesis:} Static embeddings will fail as they produce the same vector in the three cases, while models that adequately incorporate contextual cues should correctly identify the outlier sense.

\paragraph{Synonyms of homonymous senses (Exp2):} 
A word is compared with its synonym and with the synonym of its homonym, all three in different contexts
(e.g., \textit{coach}=\textit{bus}$\neq$\textit{trainer} in [1:1, 1:5, 2:2]). This test assesses if there is a bias towards one of the homonymous senses, e.g., the most frequent one 
\citep{macgregor2015sustained}.
\noindent\textbf{Hypothesis:} Models with this type of bias may fail, so as in Exp1, they should also appropriately incorporate contextual information to represent these examples.

\paragraph{Synonymy vs homonymy (Exp3):}
We compare a word to its synonym and to a homonym, all in different contexts (e.g., \textit{coach}=\textit{bus}$\neq$\textit{coach} in [1:1, 1:5, 2:1]).
Here we evaluate whether a model adequately represents both (i) synonymy in context --two word forms with the same sense in different contexts-- and (ii) homonymy --one of the former word forms having a different meaning.
\noindent\textbf{Hypothesis:} Models relying primarily on lexical knowledge are likely to represent homonyms closer than synonyms (giving rise to an incorrect output), but those integrating contextual information will be able to model the three representations correctly.

\paragraph{Synonymy (Exp4):} Two synonyms in context vs. a different word (and sense), in the same context as, at least, one of the synonyms (e.g., [2:1, 2:2, 2:3]). It assesses to what extent the context affects word representations of different word forms.
\noindent\textbf{Hypothesis:} Static embeddings may pass this test as they tend to represent type-level synonyms closely in the vector space. Highly contextualized models might be puzzled as different meanings (from different words) occur in the same context, so that the models should have an adequate trade-off between lexical and contextual knowledge.

Table~\ref{tab:eval_triples} displays the number of sentence triples for each experiment as well as the total number of triples of the dataset. To focus on the semantic knowledge encoded in the vectors --rather than on the morphosyntactic information--, we have evaluated only those triples in which the target words of the three sentences have the same POS-tag (numbers on the right).\footnote{On average, BERT-base models achieved $0.24$ higher results (\textit{Add}) when tested on all the instances (including different POS-tags) of the four experiments.} Besides, we have also carried out an evaluation on the full dataset.

\subsection{Results and discussion}

Table~\ref{tab:all_results} contains a summary of the results of each experiment in the four languages. For reasons of clarity, we include only \textit{fastText} embeddings and the best contextualized model (BERT). Results for all models and languages can be seen in Appendix~\ref{app:results}. BERT models have the best performance overall, both on the full dataset and on the selected experiments, except for Exp4 (in which the outlier shares the context at least with one of the synonyms) where the static models outperform the contextualized representations.

\begin{table*}[!ht]
\centering
\begin{tabular}{|ll|cccc|cc|c|}
\hline
\textbf{Model} & \textbf{Vec.} & \textbf{Exp1} & \textbf{Exp2} & \textbf{Exp3} & \textbf{Exp4} & \textbf{Macro} & \textbf{Micro} & \textbf{Full}\\ \hline
\multicolumn{9}{|c|}{\bf Galician} \\ \hline
\multirow{2}{*}{BERT-base} & Sent & 0.695 & 0.758 & \textbf{0.751} & 0.178 & \textbf{0.596} & \textbf{0.618} & \textbf{0.727} \\
& Cat & \textbf{0.705} & \textbf{0.799} & 0.293 & 0.422 & 0.555 & 0.513 & 0.699 \\ \hline
\multirow{3}{*}{\textit{fastText}} & Sent & 0.562 & 0.685 & 0.476 & 0.141 & 0.466 & 0.468 & 0.618 \\
& WV & 0.21\phantom{0} & 0.564 & 0\phantom{.000} & \textbf{0.526} & 0.325 & 0.286 & 0.461 \\
& Syn (3) & 0.533 & 0.658 & 0.197 & 0.185 & 0.393 & 0.362 & 0.567 \\ \hline

\multicolumn{9}{|c|}{\bf English} \\ \hline
\multirow{2}{*}{BERT-base} & Sent & 0.788 & 0.655 & 0.736 & 0.221 & 0.6\phantom{00} & 0.599 & 0.7\phantom{00} \\
& Add & \textbf{0.981} & \textbf{0.81}\phantom{0} & \textbf{0.758} & 0.441 & \textbf{0.748} & \textbf{0.732} & \textbf{0.839} \\ \hline
\multirow{3}{*}{\textit{fastText}} & Sent & 0.596 & 0.5\phantom{00} & 0.505 & 0.147 & 0.437 & 0.431 & 0.543 \\
& WV & 0.308 & 0.552 & 0.033 & \textbf{0.574} & 0.366 & 0.335 & 0.48\phantom{0} \\
& Syn (3) & 0.442 & 0.69\phantom{0} & 0.231 & 0.176 & 0.385 & 0.357 & 0.546 \\
\hline

\multicolumn{9}{|c|}{\bf Portuguese} \\ \hline
\multirow{2}{*}{BERT-base} & Sent & 0.683 & 0.432 & \textbf{0.635} & 0.22\phantom{0} & 0.493 & \textbf{0.518} & 0.564 \\
& Add & \textbf{0.854} & 0.541 & 0.378 & 0.366 & \textbf{0.535} & 0.508 & \textbf{0.67}\phantom{0} \\ \hline
\multirow{3}{*}{\textit{fastText}} & Sent & 0.61\phantom{0} & \textbf{0.622} & 0.527 & 0.171 & 0.482 & 0.487 & 0.55\phantom{0} \\
& WV & 0.024 & 0.541 & 0\phantom{.000} & \textbf{0.634} & 0.3\phantom{00} & 0.244 & 0.453 \\
& Syn (3) & 0.659 & 0.459 & 0.176 & 0.195 & 0.372 & 0.337 & 0.508 \\
\hline

\multicolumn{9}{|c|}{\bf Spanish} \\ \hline
\multirow{2}{*}{BERT-base} & Sent & 0.755 & 0.592 & \textbf{0.536} & 0.186 & 0.517 & 0.516 & 0.595 \\
& Add & \textbf{0.857} & \textbf{0.704} & 0.409 & 0.441 & \textbf{0.603} & \textbf{0.564} & \textbf{0.74}\phantom{0} \\ \hline
\multirow{3}{*}{\textit{fastText}} & Sent & 0.449 & 0.338 & 0.445 & 0.085 & 0.329 & 0.346 & 0.429 \\
& WV & 0.122 & 0.62\phantom{0} & 0.018 & \textbf{0.814} & 0.393 & 0.346 & 0.479 \\
& Syn (3) & 0.367 & 0.577 & 0.173 & 0.237 & 0.339 & 0.318 & 0.553 \\
\hline
\end{tabular}
\caption{\label{tab:all_results} Summary of the BERT and \textit{fastText} results. \textit{Macro} and \textit{Micro} refer to the macro-average and micro-average results across the four experiments, respectively. \textit{Full} are the micro-average values on the whole dataset.}
\end{table*}

\begin{figure*}[!ht]
    \centering
    \includegraphics[width=.95\textwidth]{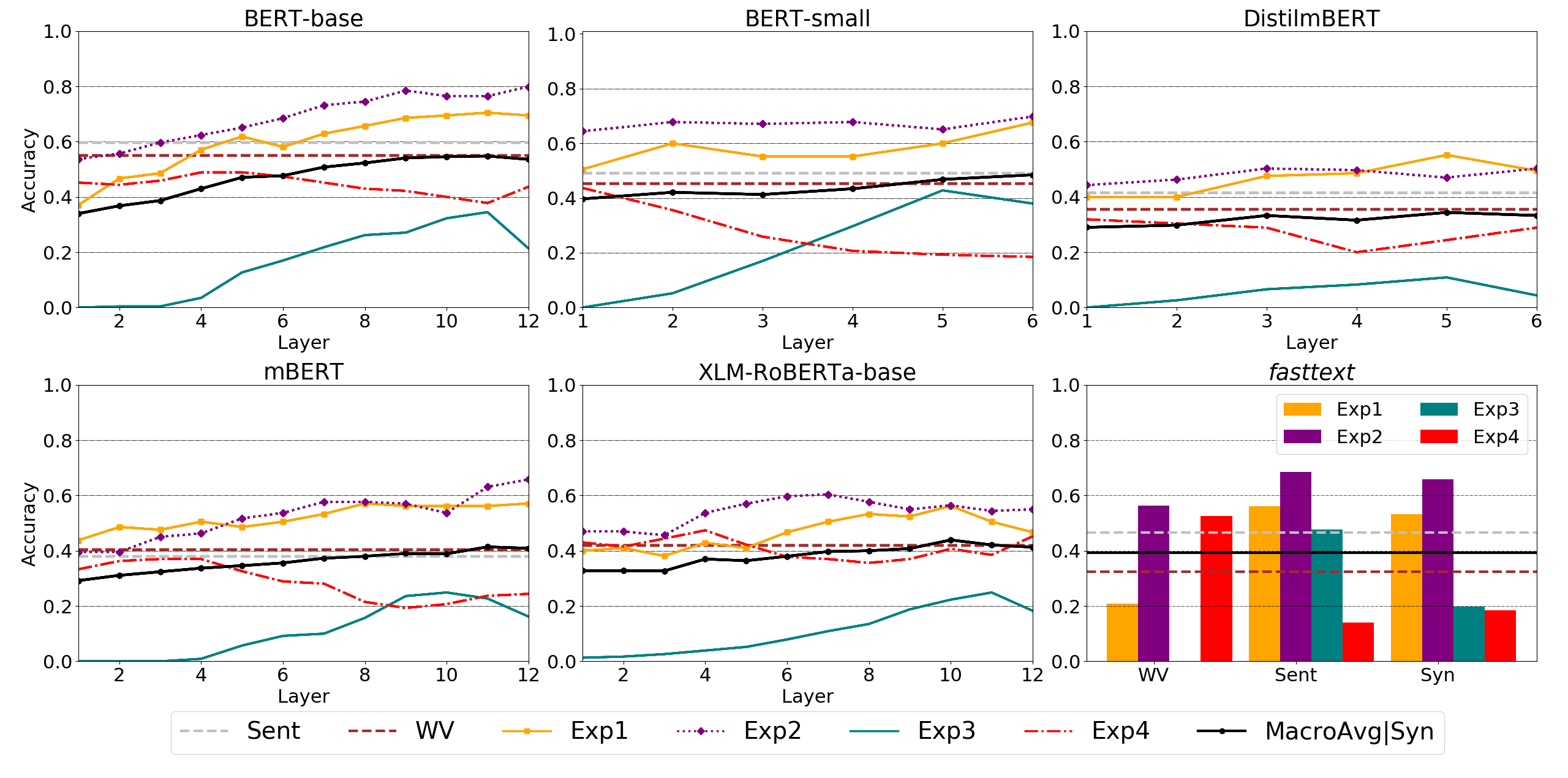}
    \caption{Results across layers and models for Galician. \textit{Sent} and \textit{WV} (dashed) are macro-average values. \textit{MacroAvg|Syn} is the macro-average per layer (Transformers) and the macro-average of the \textit{Syn} strategy (\textit{fastText}).}
    \label{fig:gl_curves}
\end{figure*}

\begin{figure*}[!ht]
    \centering
    \begin{subfigure}[b]{.49\textwidth}
        \centering
        \includegraphics[width=.95\linewidth]{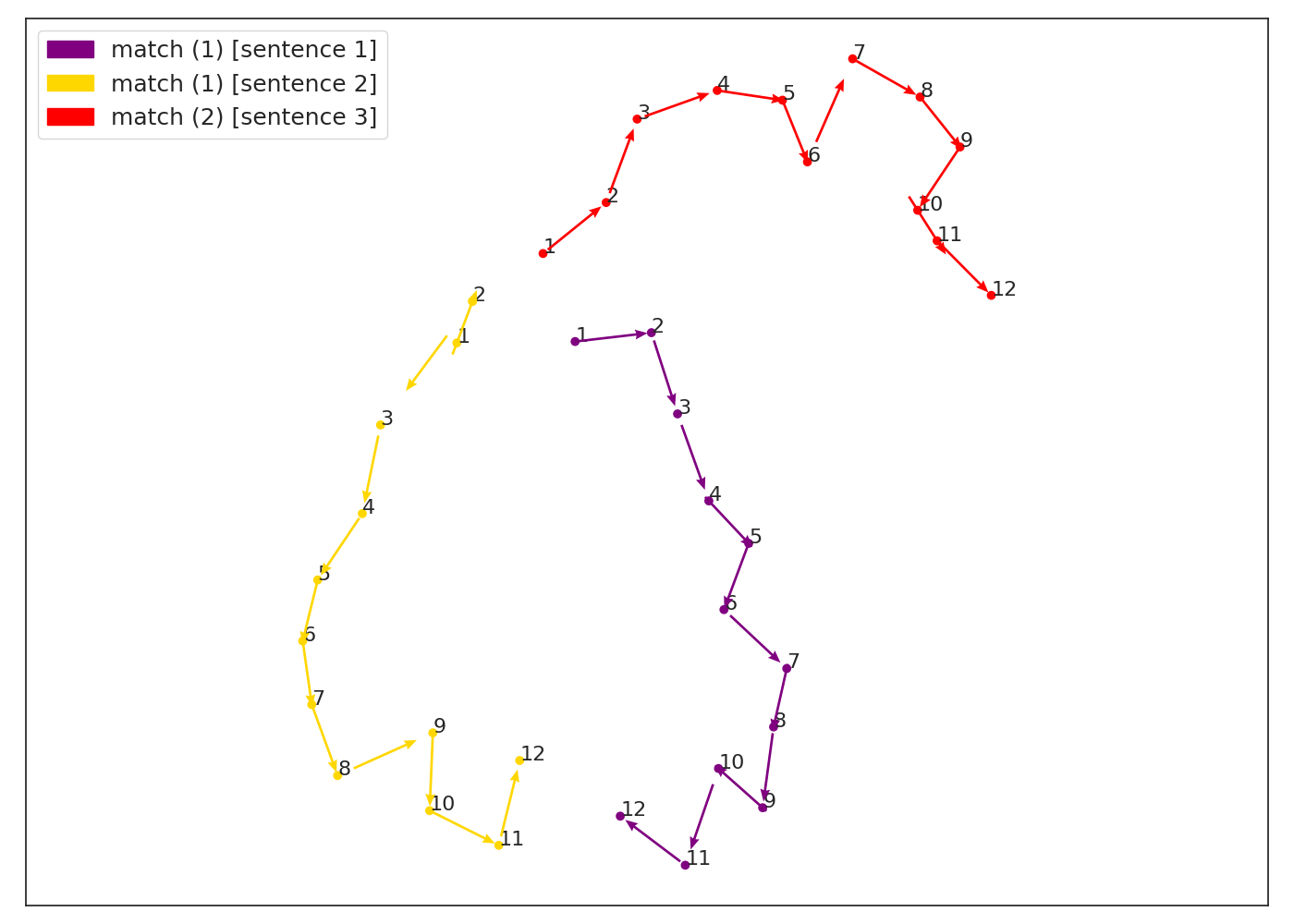}
        \caption{\label{fig:exp1}Exp1:\\
        Sentence 2: ``Chelsea have a \textit{match} with United next week.''.\\
        Sentence 3: ``You should always strike a \textit{match} away from you.''}
    \end{subfigure}
    \hfill
    \begin{subfigure}[b]{0.49\textwidth}
        \centering
        \includegraphics[width=.95\linewidth]{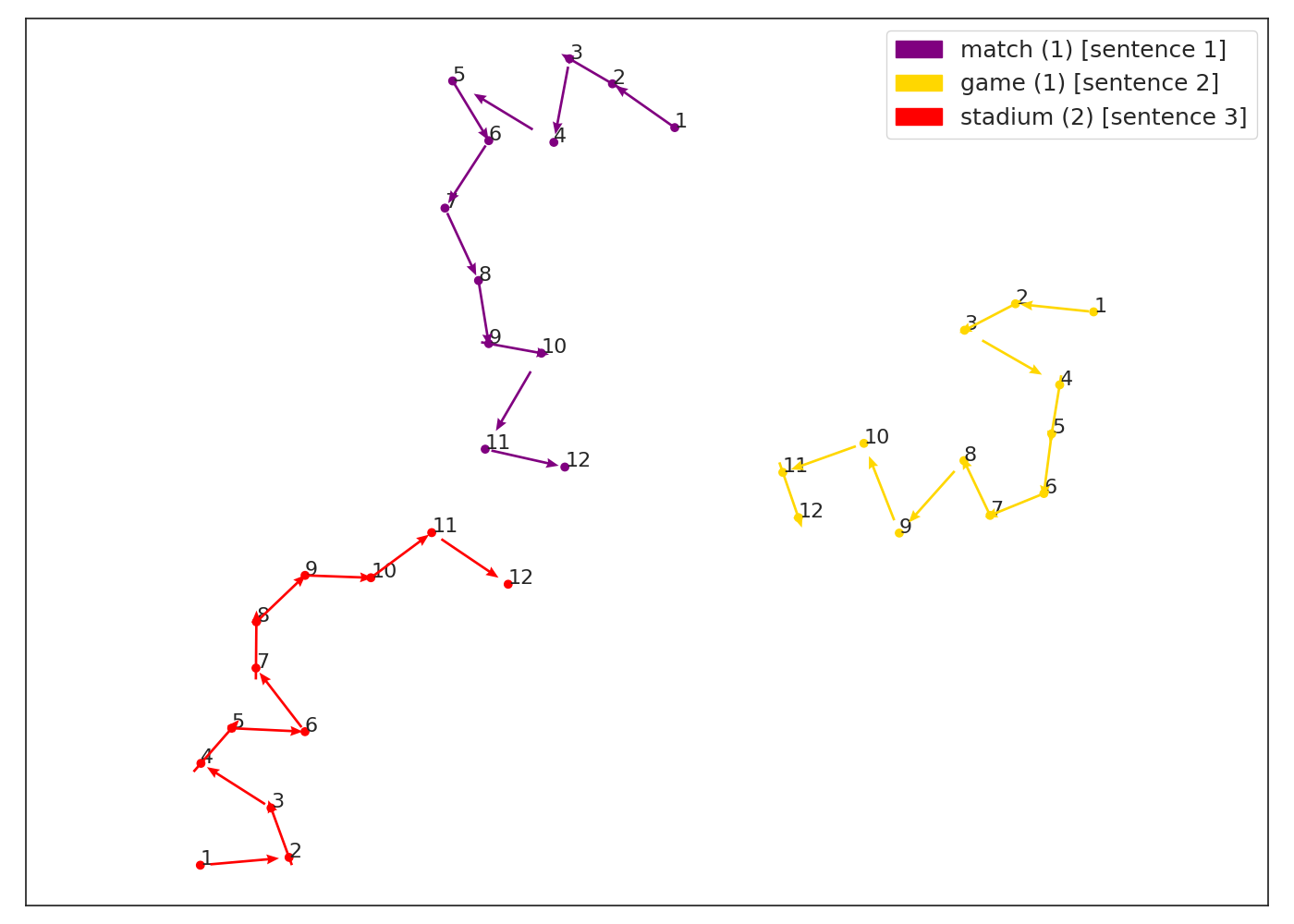}
        \caption{\label{fig:exp4}Exp4:\\
        Sentence 2: ``A \textit{game} consists of two halves lasting 45 minutes, meaning it is 90 minutes long.''.\\
        Sentence 3: ``He was watching a football \textit{stadium}.''}
    \end{subfigure}
    \caption{\label{fig:umap_exp1-4}UMAP visualizations of word contextualization across layers (1 to 12) in Exp1 and Exp4 in English (BERT-base). In both cases, sentence 1 is ``He was watching a football \textit{match}.'', and the target word in sentence 3 is the outlier.}
\end{figure*}

In Exp1 and Exp2, where the context plays a crucial role, \textit{fastText} models correctly labeled between 50\%/60\% of the examples (depending on the language and vector type, with better results for \textit{Sent} and \textit{Syn}). For BERT, the best accuracy surpasses $0.98$ (Exp1 in English), with an average across languages of $0.78$, and where word vectors outperform sentence representations. These high results and the fact that \textit{WVs} work better in general than \textit{Sent} may be indicators that Transformers are properly incorporating contextual knowledge.

Solving Exp3 requires both dealing with contextual effects and homonymy (as two words have the same form but different meaning) so that static embeddings hardly achieve $0.5$ accuracy (\textit{Sent}, with lower results for both \textit{WV} and \textit{Syn}). BERT's performance is also lower than in Exp1 and Exp2, with an average of $0.67$ and \textit{Sent} beating \textit{WVs} in most cases, indicating that the word vectors are not adequately representing the target senses.

Finally, \textit{fastText} obtains better results than BERT on Exp4 (where the outlier shares the context with at least one of the other target words), reaching $0.81$ in Spanish with an average across languages of $0.64$ (always with \textit{WVs}). BERT's best performance is $0.41$ (in two languages) with an average of $0.42$, suggesting that very similar contexts may confound the model.

To shed light on the contextualization process of Transformers, we have analyzed their performance across layers. Figure~\ref{fig:gl_curves} shows the accuracy curves (vs. the macro-average \textit{Sent} and \textit{WV} vectors of the contextualized and static embeddings) for five Transformers models on Galician, the language with the largest dataset (see Appendix~\ref{app:results} for equivalent figures for the other languages).

In Exp1 to Exp3 the best accuracies are obtained at upper layers, showing that word vectors appropriately incorporate contextual information. This is true especially for the monolingual BERT versions, as the multilingual models' representations show higher variations. Except for Galician, Exp1 has better results than Exp2, as the former primarily deals with context while the latter combines contextualization with lexical effects. In Exp3 the curves take longer to rise as initial layers rely more on lexical than on contextual information. Furthermore, except for English (which reaches $0.8$), the performance is low even in the best hidden layers ($\approx0.4$). In Exp4 (with context overlap between words with different meanings), contextualized models cannot correctly represent the word senses, being surpassed in most cases by the static embeddings.

Finally, we have observed how Transformers representations vary across the vector space. Figure~\ref{fig:umap_exp1-4} shows the UMAP visualizations \citep{McInnes2018} of the contextualization processes of Exp1 and Exp4 examples in English. In~\ref{fig:exp1}, the similar vectors of \textit{match} in layer 1 are being contextualized across layers, producing a suitable representation since layer 7. However, \ref{fig:exp4} shows how the model is not able to adequately represent \textit{match} close to its synonym \textit{game}, as the vectors seem to incorporate excessive information (or at least limited lexical knowledge) from the context. Additional visualizations in Galician can be found in Appendix~\ref{app:context}.

In sum, the experiments performed in this study allow us to observe how different models generate contextual representations. In general, our results confirm previous findings which state that Transformers models increasingly incorporate contextual information across layers. However, we have also found that this process may deteriorate the representation of the individual words, as it may be incorporating excessive contextual information, as suggested by \citet{haber-poesio-2020-assessing}.

\section{Conclusions and Further Work}
\label{sec:conclusions}
This paper has presented a systematic study of word meaning representation in context. Besides static word embeddings, we have assessed the ability of state-of-the-art monolingual and multilingual models based on the Transformers architecture to identify unambiguous cases of homonymy and synonymy. To do so, we have presented a new dataset in four linguistic varieties that allows for controlled evaluations of vector representations.

The results of our study show that, in most cases, the best contextualized models adequately identify homonyms conveying different senses in various contexts. However, as they strongly rely on the surrounding contexts, they misrepresent words having different senses in similar sentences.

In further work, we plan to extend our dataset with multiword expressions of different degrees of idiomaticity and to include less transparent --but still unambiguous-- contexts of homonymy. Finally, we also plan to systematically explore how multilingual models represent homonymy and synonymy in cross-lingual scenarios.

\section*{Acknowledgments}
We would like to thank the anonymous reviewers for their valuable comments, and NVIDIA Corporation for the donation of a Titan Xp GPU. This research is funded by a \textit{Ramón y Cajal} grant (RYC2019-028473-I) and by the Galician Government (ERDF 2014-2020: Call ED431G 2019/04).

\bibliography{anthology,acl2021}
\bibliographystyle{acl_natbib}

\clearpage

\appendix

\onecolumn
\section*{Appendices}

\section{Complete results}
\label{app:results}
\begin{figure*}[!h]
    \centering
    \includegraphics[width=.82\textwidth, frame]{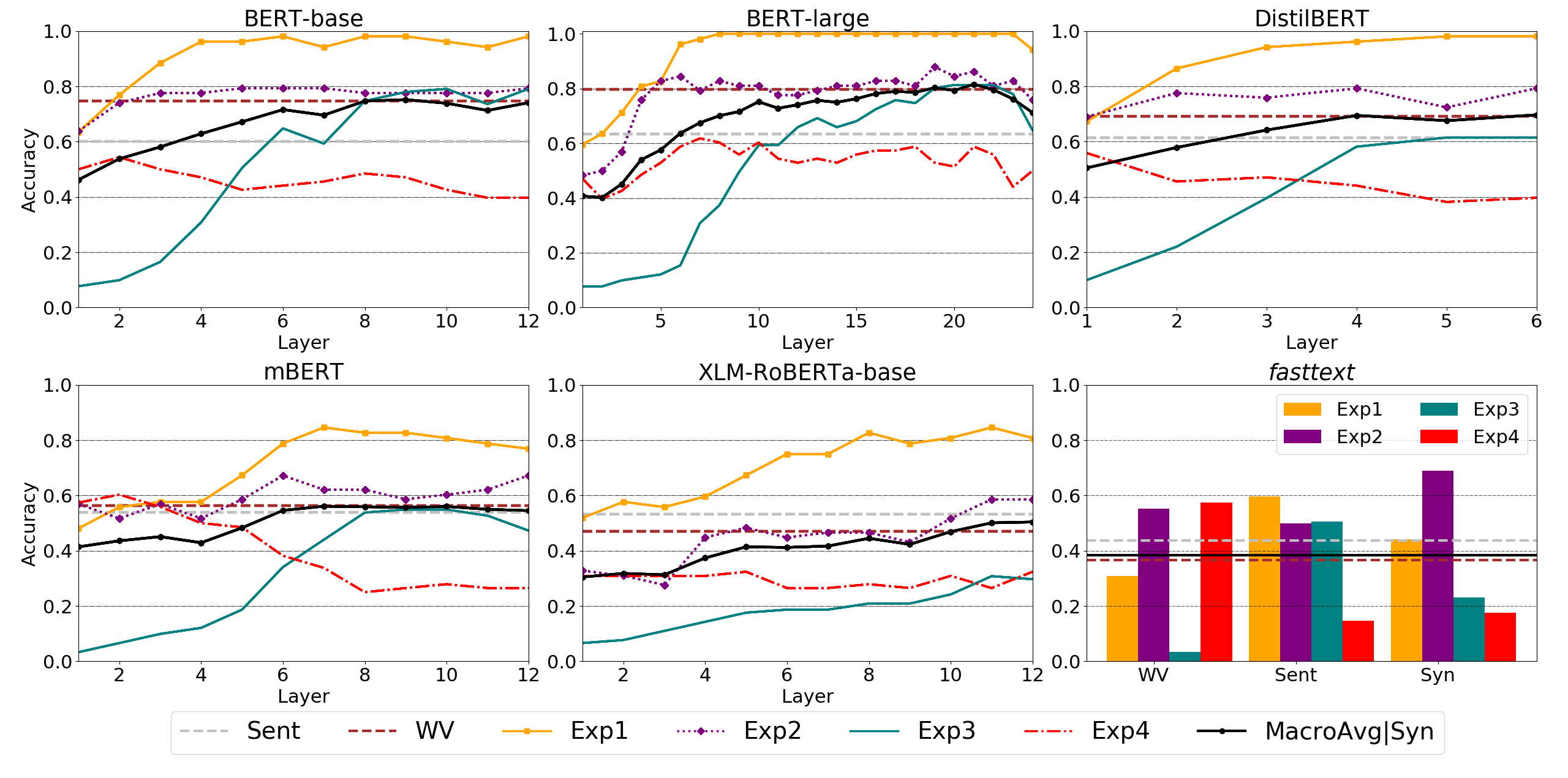}\\
    \vfill
    \includegraphics[width=.82\textwidth, frame]{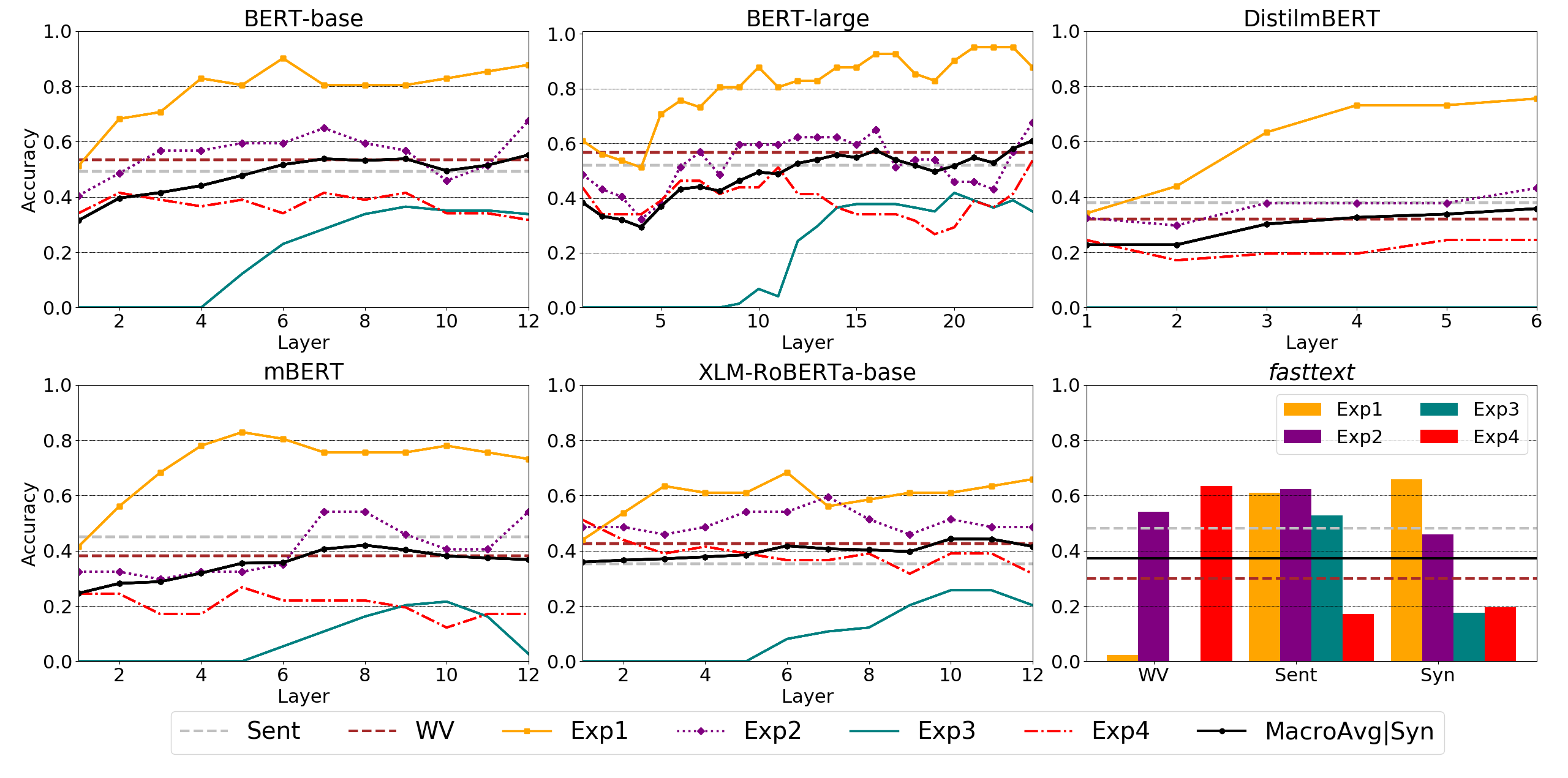}\\
    \vfill
    \includegraphics[width=.82\textwidth, frame]{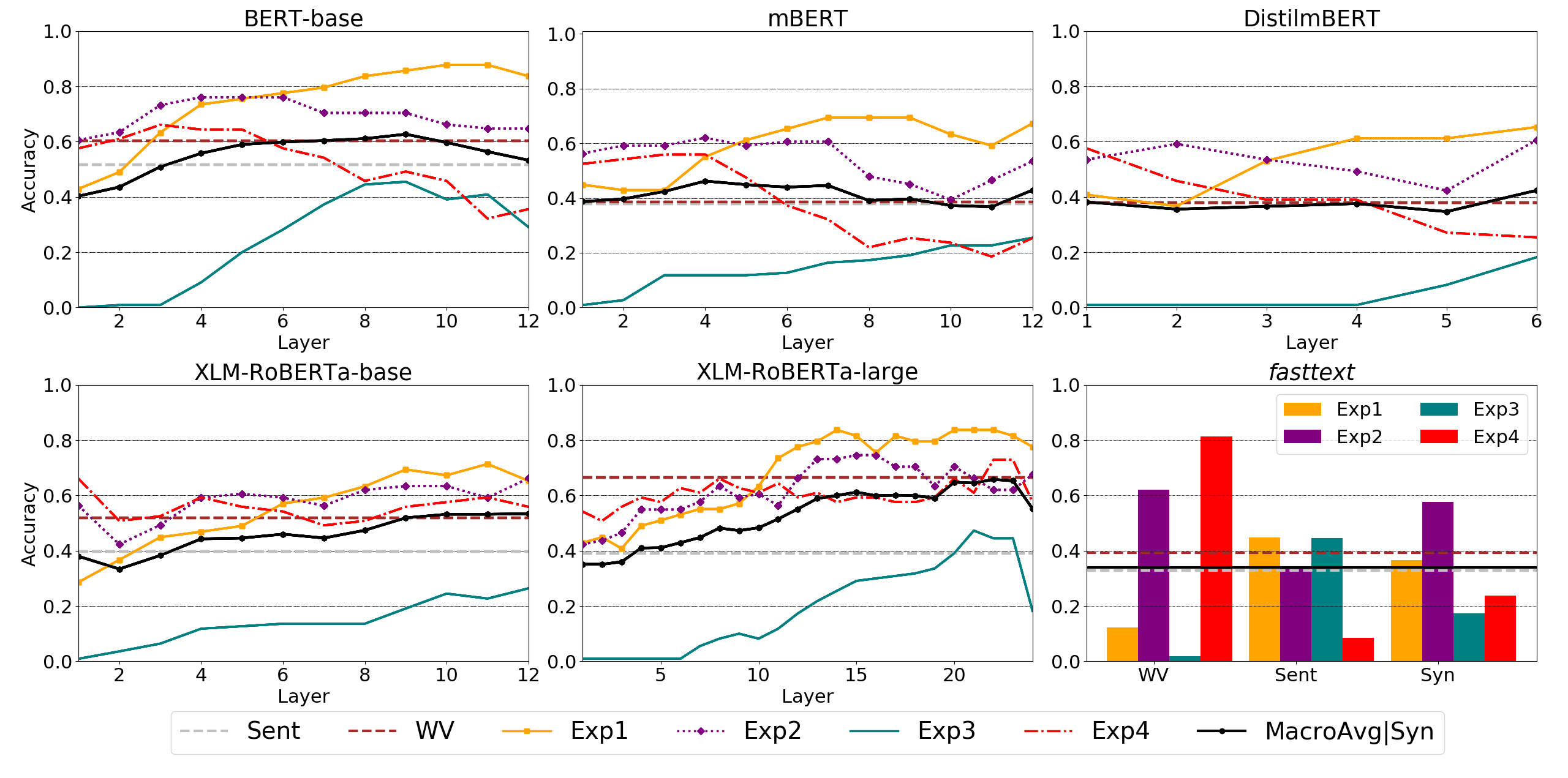}
    \caption{Results across layers and models for English (top), Portuguese (middle), and Spanish (bottom). \textit{Sent} and \textit{WV} (dashed) are macro-average values. \textit{MacroAvg|Syn} is the macro-average per layer (Transformers) and the macro-average of the \textit{Syn} strategy (\textit{fastText}).}
    \label{fig:all_curves}
\end{figure*}

Figure~\ref{fig:all_curves} and Table~\ref{tab:full_results} include the results for all languages and models. We also include \textit{large} variants (BERT and XLM-RoBERTa) when available. For static embeddings, we report results for the best \textit{Syn} setting, which combines up to three syntactically related words with the target word (see Appendix~\ref{app:syn}).

\begin{sidewaystable}
\centering
{\fontsize{9}{12}\selectfont{
\setlength{\tabcolsep}{2.75pt}
\begin{tabular}{|ll|cccc|cc|c||cccc|cc|c||cccc|cc|c||cccc|cc|c|}
\hline

\textbf{Model} & \textbf{Vec.} & \textbf{E1} & \textbf{E2} & \textbf{E3} & \textbf{E4} & \textbf{Ma} & \textbf{Mi} & \textbf{F}
& \textbf{E1} & \textbf{E2} & \textbf{E3} & \textbf{E4} & \textbf{Ma} & \textbf{Mi} & \textbf{F}
& \textbf{E1} & \textbf{E2} & \textbf{E3} & \textbf{E4} & \textbf{Ma} & \textbf{Mi} & \textbf{F}
& \textbf{E1} & \textbf{E2} & \textbf{E3} & \textbf{E4} & \textbf{Ma} & \textbf{Mi} & \textbf{F} \\ \hline

\multicolumn{2}{|c}{} & \multicolumn{7}{|c||}{\bf Galician}
& \multicolumn{7}{|c||}{\bf Portuguese}
& \multicolumn{7}{|c||}{\bf Spanish}
& \multicolumn{7}{|c|}{\bf English} \\ \hline

\multirow{3}{*}{BERT} & Sent & 0.7\phantom{0} & 0.76 & \textbf{0.75} & 0.18 & \textbf{0.6}\phantom{0} & \textbf{0.62}
& \textbf{0.73}
& 0.68 & 0.43 & 0.64 & 0.22 & 0.49 & 0.52 & 0.56
& 0.76 & 0.59 & \textbf{0.54} & 0.19 & 0.52 & 0.52 & 0.6\phantom{0}
& 0.79 & 0.66 & 0.74 & 0.22 & 0.6\phantom{0} & 0.6\phantom{0} & 0.7\phantom{0} \\

& Cat & \textbf{0.71} & \textbf{0.8}\phantom{0} & 0.29 & 0.42 & 0.56 & 0.51 & 0.7\phantom{0}
& 0.85 & 0.51 & 0.38 & 0.37 & 0.53 & 0.5\phantom{0} & 0.66
& \textbf{0.86} & \textbf{0.7}\phantom{0} & 0.41 & 0.44 & 0.6\phantom{0} & 0.56 & 0.74
& 0.96 & \textbf{0.83} & 0.76 & 0.43 & 0.74 & 0.73 & 0.84 \\

& Add & 0.7\phantom{0} & \textbf{0.8}\phantom{0} & 0.28 & 0.42 & 0.55 & 0.51 & 0.7\phantom{0}
& 0.85 & 0.54 & 0.38 & 0.37 & 0.54 & 0.51 & 0.67
& \textbf{0.86} & \textbf{0.7}\phantom{0} & 0.41 & 0.44 & 0.6\phantom{0} & 0.56 & 0.74
& 0.98 & 0.81 & 0.76 & 0.44 & 0.75 & 0.73 & 0.84 \\ \hline

\multirow{3}{*}{BERT\textsubscript{2}} & Sent & 0.61 & 0.6\phantom{0} & 0.59 & 0.16 & 0.49 & 0.5\phantom{0} & 0.64
& 0.68 & 0.49 & \textbf{0.69} & 0.22 & 0.52 & \textbf{0.55} & 0.6\phantom{0} 
&--&--&--&--&--&--&--
& 0.89 & 0.59 & \textbf{0.8}\phantom{0} & 0.27 & 0.63 & 0.64 & 0.7\phantom{0} \\

& Cat & 0.62 & 0.71 & 0.3\phantom{0} & 0.2\phantom{0} & 0.46 & 0.43 & 0.65
& \textbf{0.95} & 0.51 & 0.38 & 0.46 & \textbf{0.58} & 0.54 & \textbf{0.68} 
&--&--&--&--&--&--&--
& \textbf{1}\phantom{.00} & 0.81 & 0.78 & \textbf{0.57} & 0.79 & \textbf{0.78} & \textbf{0.87} \\

& Add & 0.61 & 0.71 & 0.29 & 0.2\phantom{0} & 0.45 & 0.43 & 0.65
& \textbf{0.95} & 0.49 & 0.37 & 0.46 & 0.57 & 0.53 & \textbf{0.68} 
&--&--&--&--&--&--&--
& \textbf{1}\phantom{.00} & 0.81 & \textbf{0.8}\phantom{0} & \textbf{0.57} & \textbf{0.8}\phantom{0} & \textbf{0.78} & \textbf{0.87} \\ \hline

\multirow{3}{*}{mBERT} & Sent & 0.48 & 0.4\phantom{0} & 0.49 & 0.16 & 0.38 & 0.39 & 0.53
& 0.63 & 0.43 & 0.57 & 0.17 & 0.45 & 0.47 & 0.54
& 0.51 & 0.41 & 0.41 & 0.19 & 0.38 & 0.38 & 0.5\phantom{0}
& 0.65 & 0.57 & 0.77 & 0.16 & 0.54 & 0.55 & 0.61 \\

& Cat & 0.57 & 0.61 & 0.23 & 0.22 & 0.41 & 0.38 & 0.62
& 0.73 & 0.46 & 0.16 & 0.15 & 0.38 & 0.34 & 0.54
& 0.61 & 0.45 & 0.23 & 0.24 & 0.38 & 0.35 & 0.63
& 0.83 & 0.62 & 0.53 & 0.27 & 0.56 & 0.54 & 0.73 \\

& Add & 0.57 & 0.62 & 0.21 & 0.22 & 0.4\phantom{0} & 0.37 & 0.61
& 0.73 & 0.49 & 0.14 & 0.17 & 0.38 & 0.34 & 0.55
& 0.63 & 0.44 & 0.22 & 0.25 & 0.39 & 0.35 & 0.63
& 0.83 & 0.62 & 0.54 & 0.27 & 0.56 & 0.54 & 0.73 \\ \hline

\multirow{3}{*}{XLM-b} & Sent & 0.52 & 0.51 & 0.49 & 0.16 & 0.42 & 0.43 & 0.54
& 0.51 & 0.3\phantom{0} & 0.41 & 0.2\phantom{0} & 0.35 & 0.36 & 0.45
& 0.51 & 0.44 & 0.46 & 0.19 & 0.4\phantom{0} & 0.41 & 0.51
& 0.6\phantom{0} & 0.62 & 0.69 & 0.22 & 0.53 & 0.54 & 0.63 \\

& Cat & 0.56 & 0.54 & 0.22 & 0.38 & 0.42 & 0.39 & 0.56
& 0.63 & 0.46 & 0.24 & 0.34 & 0.42 & 0.39 & 0.61
& 0.67 & 0.62 & 0.23 & 0.54 & 0.52 & 0.46 & 0.69
& 0.83 & 0.59 & 0.23 & 0.27 & 0.48 & 0.43 & 0.69 \\

& Add & 0.55 & 0.54 & 0.2\phantom{0} & 0.39 & 0.42 & 0.38 & 0.56
& 0.63 & 0.51 & 0.22 & 0.34 & 0.43 & 0.39 & 0.61
& 0.67 & 0.62 & 0.23 & 0.56 & 0.52 & 0.47 & 0.69
& 0.81 & 0.55 & 0.23 & 0.29 & 0.47 & 0.43 & 0.68 \\ \hline

\multirow{3}{*}{XLM-l} & Sent & 0.42 & 0.34 & 0.42 & 0.16 & 0.33 & 0.34 & 0.44
& 0.49 & 0.43 & 0.35 & 0.15 & 0.35 & 0.35 & 0.44
& 0.49 & 0.48 & 0.39 & 0.2\phantom{0} & 0.39 & 0.39 & 0.47
& 0.54 & 0.5\phantom{0} & 0.55 & 0.22 & 0.45 & 0.45 & 0.58 \\

& Cat & 0.48 & 0.5\phantom{0} & 0.22 & 0.42 & 0.4\phantom{0} & 0.37 & 0.49
& 0.73 & 0.49 & 0.39 & 0.32 & 0.48 & 0.47 & 0.58
& 0.84 & 0.63 & 0.46 & 0.71 & 0.66 & 0.62 & 0.76
& 0.71 & 0.6\phantom{0} & 0.49 & 0.41 & 0.55 & 0.54 & 0.62 \\

& Add & 0.46 & 0.51 & 0.2\phantom{0} & 0.43 & 0.4\phantom{0} & 0.37 & 0.5\phantom{0}
& 0.73 & 0.51 & 0.38 & 0.32 & 0.49 & 0.47 & 0.58
& 0.84 & 0.66 & 0.46 & 0.71 & \textbf{0.67} & \textbf{0.62} & \textbf{0.77}
& 0.73 & 0.6\phantom{0} & 0.51 & 0.46 & 0.57 & 0.56 & 0.64 \\ \hline

\multirow{3}{*}{DmBERT} & Sent & 0.51 & 0.49 & 0.5\phantom{0} & 0.16 & 0.42 & 0.43 & 0.57
& 0.51 & 0.43 & 0.47 & 0.1\phantom{0} & 0.38 & 0.39 & 0.5\phantom{0}
& 0.51 & 0.44 & 0.45 & 0.12 & 0.38 & 0.39 & 0.51
& 0.67 & 0.55 & 0.79 & 0.24 & 0.56 & 0.58 & 0.63 \\

& Cat & 0.52 & 0.52 & 0.07 & 0.24 & 0.34 & 0.29 & 0.51
& 0.68 & 0.32 & 0\phantom{.00} & 0.22 & 0.31 & 0.25 & 0.47
& 0.61 & 0.49 & 0.01 & 0.34 & 0.36 & 0.3\phantom{0} & 0.53
& 0.69 & 0.52 & 0.24 & 0.28 & 0.43 & 0.4\phantom{0} & 0.63 \\

& Add & 0.54 & 0.56 & 0.07 & 0.26 & 0.36 & 0.31 & 0.51
& 0.71 & 0.35 & 0\phantom{.00} & 0.22 & 0.32 & 0.26 & 0.47
& 0.61 & 0.52 & 0.01 & 0.37 & 0.38 & 0.31 & 0.54
& 0.69 & 0.53 & 0.21 & 0.37 & 0.45 & 0.41 & 0.63 \\ \hline

\multirow{3}{*}{\textit{fastT}} & Sent & 0.56 & 0.69 & 0.48 & 0.14 & 0.47 & 0.47 & 0.62
& 0.61 & \textbf{0.62} & 0.53 & 0.17 & 0.48 & 0.49 & 0.55
& 0.45 & 0.34 & 0.45 & 0.09 & 0.33 & 0.35 & 0.43
& 0.6\phantom{0} & 0.5\phantom{0} & 0.51 & 0.15 & 0.44 & 0.43 & 0.54 \\

& WV & 0.21 & 0.56 & 0\phantom{.00} & \textbf{0.53} & 0.33 & 0.29 & 0.46
& 0.02 & 0.54 & 0\phantom{.00} & \textbf{0.63} & 0.3\phantom{0} & 0.24 & 0.45
& 0.12 & 0.62 & 0.02 & \textbf{0.81} & 0.39 & 0.35 & 0.48
& 0.31 & 0.55 & 0.03 & \textbf{0.57} & 0.37 & 0.34 & 0.48 \\

& Syn (3) & 0.53 & 0.66 & 0.2\phantom{0} & 0.19 & 0.39 & 0.36 & 0.57
& 0.66 & 0.46 & 0.18 & 0.2\phantom{0} & 0.37 & 0.34 & 0.51
& 0.37 & 0.58 & 0.17 & 0.24 & 0.34 & 0.32 & 0.55
& 0.44 & 0.69 & 0.23 & 0.18 & 0.39 & 0.36 & 0.55 \\

\hline
\end{tabular}
}}
\caption{\label{tab:full_results} Complete results for the four languages. BERT are BERT-Base models, and BERT\textsubscript{2} refers to a second BERT model for each language (small for Galician, and large for Portuguese and English). XLM-b and XLM-l are XLM-RoBERTa base and large models, respectively. DmBERT is the multilingual version of DistilBERT, and \textit{fastT} the \textit{fastText} embeddings. \textit{Ma} and \textit{Mi} refer to the macro-average and micro-average results across the four experiments, respectively. \textit{F} are the micro-average values on the whole dataset.}

\end{sidewaystable}

\clearpage
\section{Contextualization process}
\label{app:context}

\begin{figure*}[!ht]
    \centering
    \begin{subfigure}[b]{.49\textwidth}
        \centering
        \includegraphics[width=.8\linewidth]{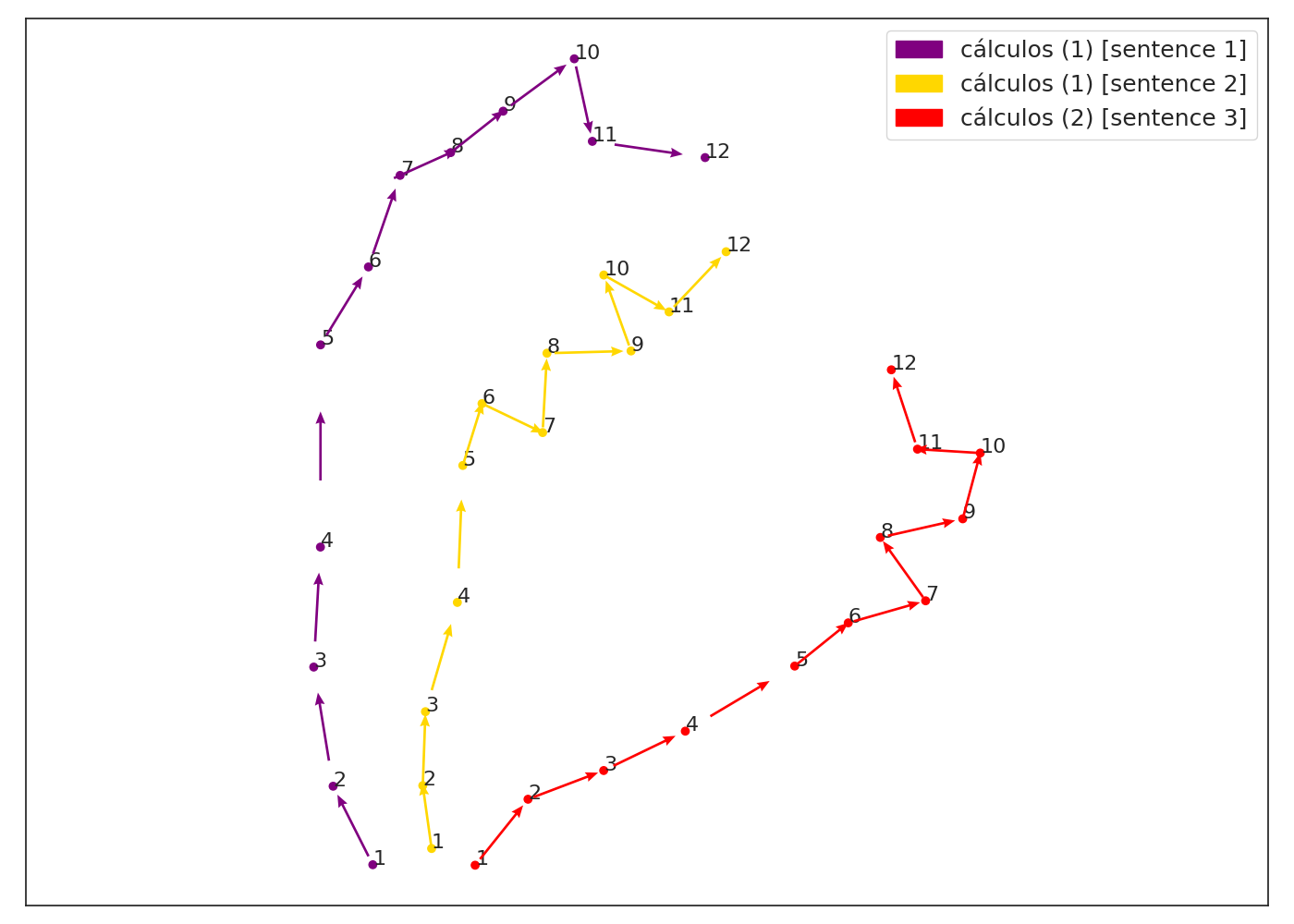}
        \caption{\label{ok1}Sent. 1: ``Ten que haber algún erro nos \textit{cálculos} porque o resultado non é correcto.''\\Sent. 2: ``Segundo os meus \textit{cálculos} acabaremos en tres días.''\\Sent. 3: ``Tivo varios \textit{cálculos} biliares.''}
    \end{subfigure}
    \hfill
    \begin{subfigure}[b]{0.49\textwidth}
        \centering
        \includegraphics[width=.8\linewidth]{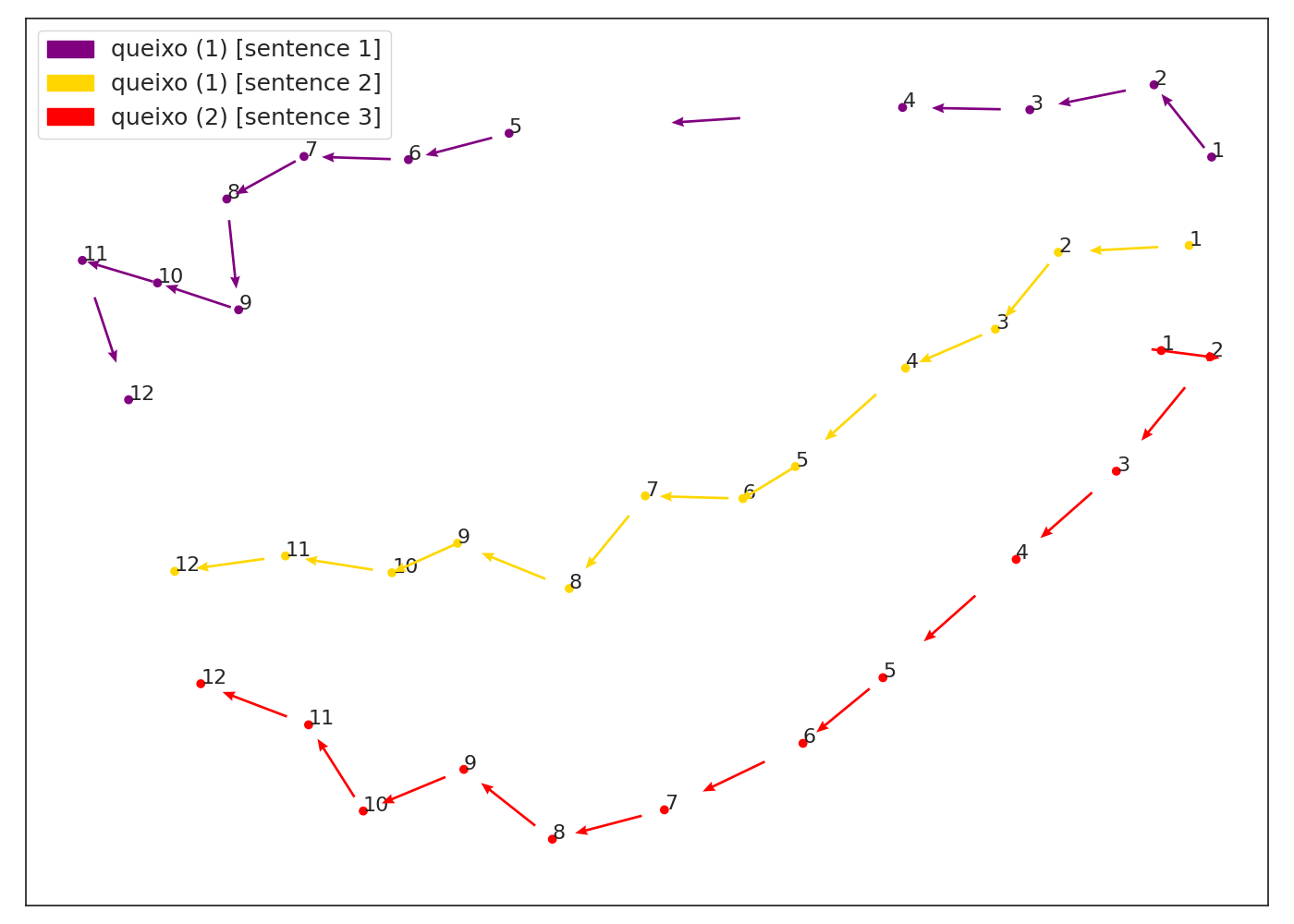}
        \caption{\label{no1}Sent. 1: ``De sobremesa tomou \textit{queixo} con marmelo.''\\Sentence 2: ``Fomos a unhas xornadas gastronómicas do \textit{queixo}.''\\Sentence 3: ``Achegouse a ela e pasoulle a man polo \textit{queixo}.''}
    \end{subfigure}

    \vfill
    
    \centering
    \begin{subfigure}[b]{.49\textwidth}
        \centering
        \includegraphics[width=.8\linewidth]{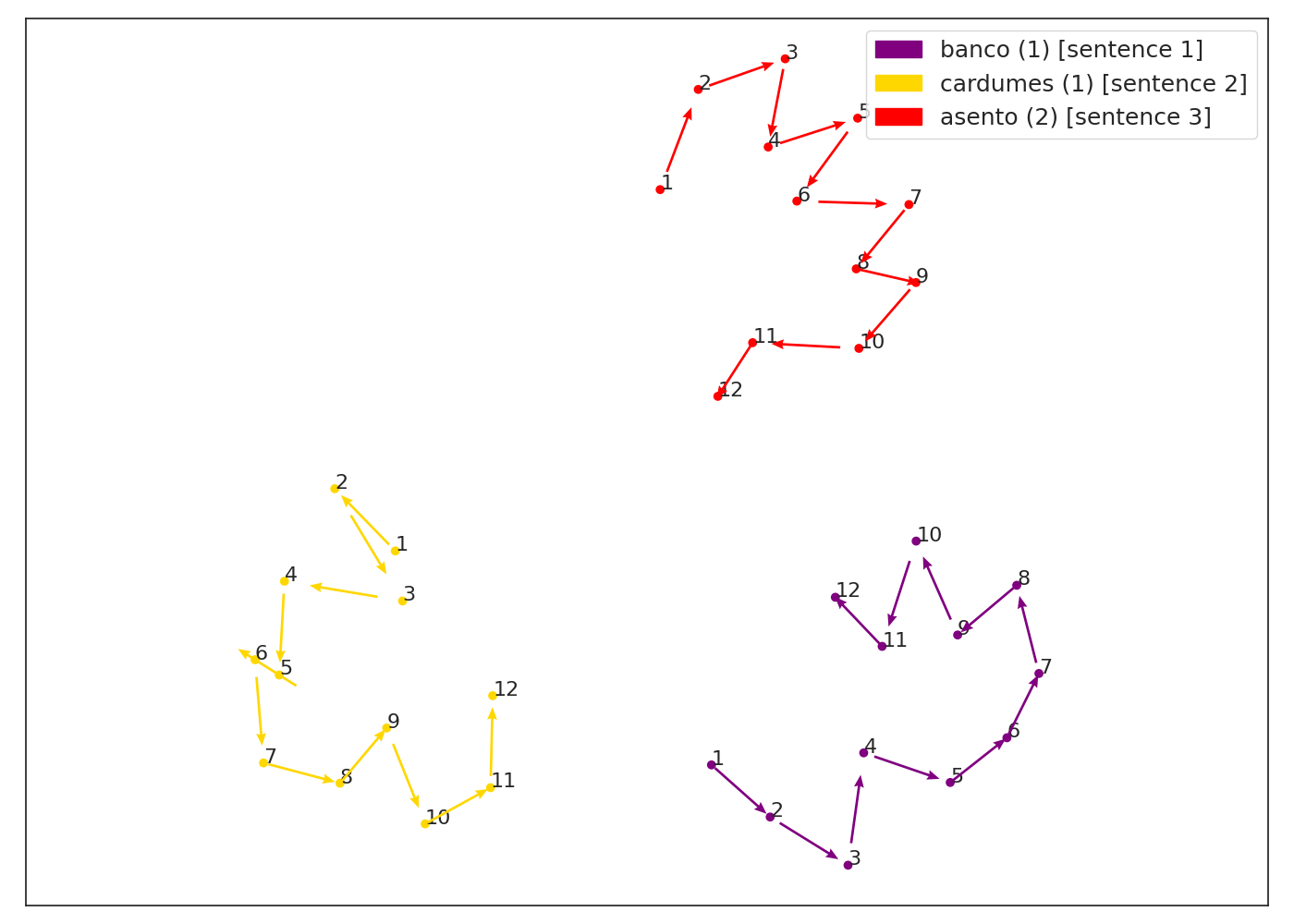}
        \caption{\label{sf2a}Sentence 1: ``Eran tantos que parecían un \textit{banco} de xurelos.''\\Sent.2: ``Desde a rocha víanse pequenos \textit{cardumes} de robaliza.''\\Sentence 3: ``Este \textit{asento} de pedra é algo incómodo.''}
    \end{subfigure}
    \hfill
    \begin{subfigure}[b]{0.49\textwidth}
        \centering
        \includegraphics[width=.8\linewidth]{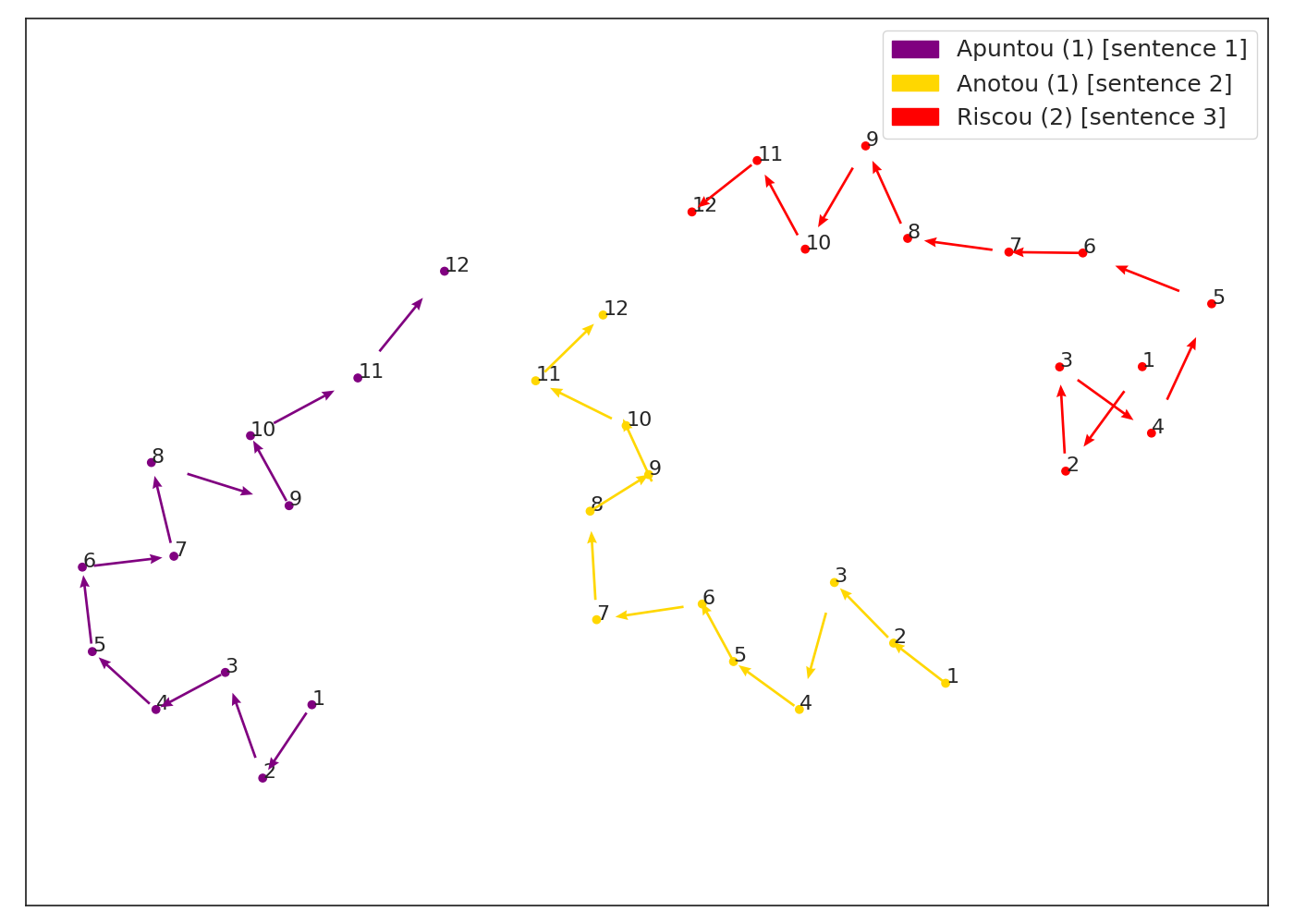}
        \caption{\label{sf2b}Sent.1: ``\textit{Apuntou} todos os números de teléfono na axenda.''\\Sentence 2: ``\textit{Anotou} todos os números de teléfono na axenda.''\\Sentence 3: ``\textit{Riscou} todos os números de teléfono na axenda.''.}
    \end{subfigure}
    
    \vfill
    
    \centering
    \begin{subfigure}[b]{.49\textwidth}
        \centering
        \includegraphics[width=.8\linewidth]{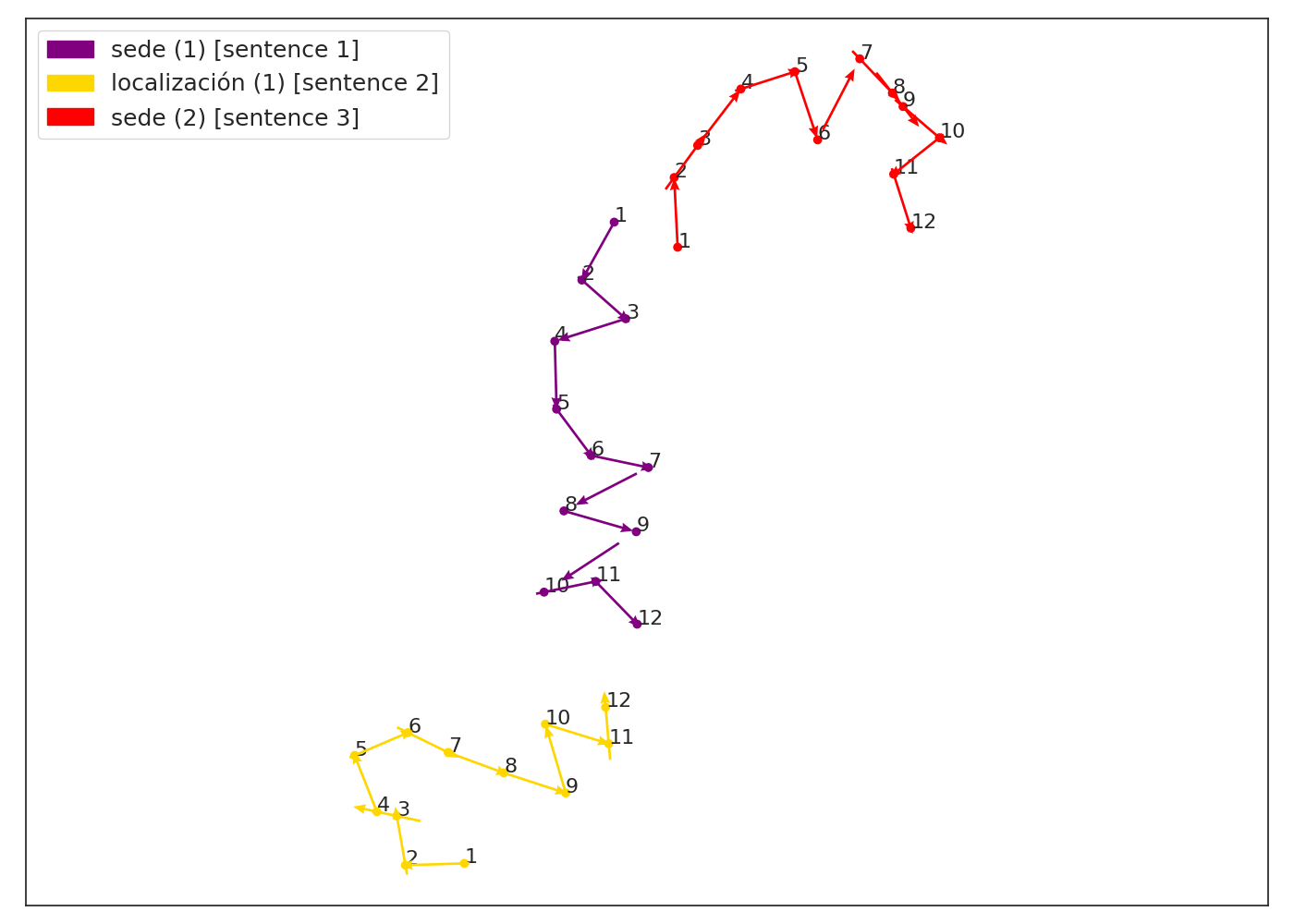}
        \caption{\label{sf3a}Sent. 1: ``Vai ter lugar a elección da próxima \textit{sede} dos Xogos Olímpicos.''\\Sent. 2: ``A \textit{localización} do evento será decidida esta semana.''\\Sent. 3: ``Vou á fonte por auga, que teño \textit{sede}.''}
    \end{subfigure}
    \hfill
    \begin{subfigure}[b]{0.49\textwidth}
        \centering
        \includegraphics[width=.8\linewidth]{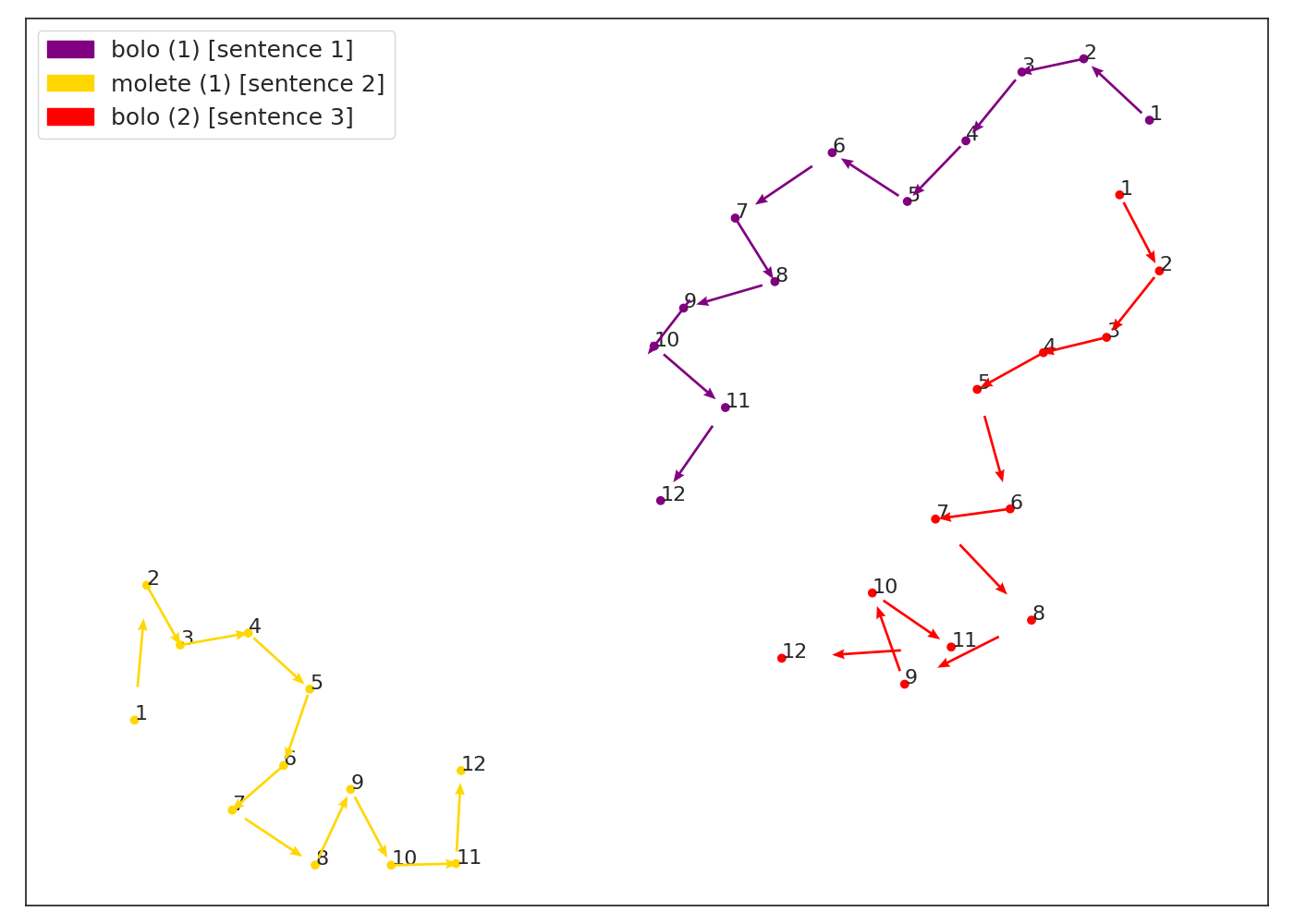}
        \caption{\label{sf3b}Sentence 1: ``Encántalle comer o \textit{bolo} de pan antes da sopa.''\\Sentence 2: ``O \textit{molete} tiña a codia un pouco dura.''\\Sentence 3: ``Para atraeren as robalizas iscaban \textit{bolo} vivo.''\\}
    \end{subfigure}
    \caption{Examples in Galician using BERT-base (English translations of the sentences in Appendix~\ref{app:translations}).\newline
    First row shows examples of Ex1. In Figure~\ref{ok1} \textit{cálculos} is correctly contextualized since layer 3. In Figure~\ref{no1}, the outlier sense of \textit{queixo} is not correctly contextualized in any layer.\newline
    Second row shows examples of Exp2 (\ref{sf2a}) and Exp4 (\ref{sf2b}). In Figure~\ref{sf2a}, the synonymys \textit{banco} and \textit{cardume} are closer to the outlier \textit{asento} in layer 1 (and from 4 to 7), but the contextualization process is not able to correctly represent the senses in the vector space. In Figure~\ref{sf2b}, the result is correct from layer 7 to 11, but in general the representations of words in similar sentences point towards a similar region.\newline
    Third row incudes examples of Exp3. In Figure~\ref{sf3a}, the occurrences of the homonym \textit{sede} are correctly contextualized as the one in the first sentence approaches its synonym \textit{localización} in upper layers. The equivalent example of Figure~\ref{sf3b} is not adequately solved by the model, as both senses of \textit{bolo} are notoriously distanct from \textit{molete}, synonym of the first homonymous sense.
    }\label{fig:umap_gl}
\end{figure*}

\twocolumn

\section{Galician models}
\label{app:models}

\paragraph{Training corpus:}
We combined the SLI GalWeb \citep{agerri-etal-2018-developing}, CC-100 \citep{wenzek-etal-2020-ccnet}, the Galician Wikipedia (April 2020 dump), and other news corpora crawled from the web. Following \citet{raffel2020exploring}, sentences with a high ratio of punctuation and symbols, and duplicates were removed. 
The final corpus has 555M words (633M tokens tokenized with FreeLing \citep{padro-stanilovsky-2012-freeling,garcia2010}). The corpus was divided into 90\%/10\% splits for train and development.

\paragraph{\textit{fastText} model:}
We trained a \textit{fastText} skip-gram model for 15 iterations with 300 dimensions, window size of 5, negative sampling of 25, and a minimum word frequency of 5. We used the same 90\% split used to train the BERT models, but with automatic tokenization ($\approx$ 600M tokens).

\paragraph{BERT models:}
We used the 90\% train split of the corpus (with the original tokenization) to train two BERT models, with 6 and 12 layers:

\subparagraph{BERT-small (6 layers):} This model has been trained from scratch using a vocabulary of 52,000 (sub-)words and a batch size of 208. It has been training during 1M steps ($\approx20$ epochs) in 14 days.

\subparagraph{BERT-base (12 layers):} Following \citet{yu2019adaptation}, we initialized the model from the official pre-trained mBERT, therefore having the same vocabulary size (119,547). We trained it on the Galician corpus during 600k steps ($\approx13$ epochs in 28 days) with a batch size of 198.

Both models were trained with the \textit{Transformers} library \citep{wolf-etal-2020-transformers} 
on a single NVIDIA Titan XP GPU (12GB), a block size of 128, a learning rate of 0.0001, a masked language modeling (MLM) probability of 0.15, and a weight decay of 0.01. They have been trained only with the MLM objective.

\section{Syntax (\textit{Syn} method)}
\label{app:syn}
To get the heads and dependents of each target word we have used the following hierarchies: For nouns: $HeadVerb$ (the head verb, if any)$>DepVerb$ (dependents of the head verb with one of the following relations: \textit{obj}, \textit{nmod}, \textit{obl})$>DepAdj$ (a dependent adjective)$>DepNoun$ (a dependent noun). For verbs: $Head$ (only if it is a verb or a noun)$>Obj$ (its direct object, if any)$>Arg$ (a dependent with one of these relations: \textit{nsubj}, \textit{nmod}, \textit{obl}). Using these hierarchies we have evaluated representations built by adding from 1 to 4 vectors to the one of each target word. As shown in Table~\ref{tab:full_results}, combining 3 syntactically related words to the target one obtains the best results.

For the experiments, we have parsed the datasets using the 2.5 \textit{Universal Dependencies} models provided by UDPipe \cite{straka-etal-2019-udpipe}.

\section{English translations (Figure~\ref{fig:umap_gl})}\label{app:translations}
Figure~\ref{ok1}, sentence 1: ``There must be some error in the \textit{calculations} because the result is incorrect''. Sentence 2: ``According to my \textit{calculations} we will finish in three days''. Sentence 3: ``[He/she] had several \textit{gallstones}''.

Figure~\ref{no1}, sentence 1: ``For dessert [he/she] ate \textit{cheese} with quince''. Sentence 2: ``We went to a \textit{cheese} gastronomy days''. Sentence 3: ``[He/She] approached her and ran his hand over her \textit{chin}''.

Figure~\ref{sf2a}, sentence 1: ``They were so many that they looked like a \textit{school} of mackerel''. Sentence 2: ``From the rock small \textit{shoals} of sea bass could be seen''. Sentence 3: ``This stone \textit{seat} is somewhat uncomfortable''.

Figure~\ref{sf2b}, sentences 1 and 2: ``[He/She] \textit{wrote down} all the phone numbers on the phone book.'' Sentence 3: ``[He/She] \textit{crossed out} all the phone numbers on the phone book''.

Figure~\ref{sf3a}, sentence 1: ``The choice of the next \textit{venue} for the Olympics will take place''. Sentence 2: ``The \textit{location} of the event will be decided this week''. Sentence 3: ``I'll get water from the spring, I am \textit{thirsty}''.

Figure~\ref{sf3b}, sentence 1: ``[He/She] loves to eat the \textit{bread cake} before soup''. Sentence 2: ``The \textit{bread} had a slightly hard crust''. Sentence 3: ``They used live \textit{sand lance} to attrack sea bass''.

\end{document}